\documentclass[sigconf]{acmart}
\AtBeginDocument{%
  }

\begin{document}

\pagestyle{plain}

\title{GRADE: Personalized Multi-Task Fusion via Group-relative Reinforcement Learning with Adaptive Dirichlet Exploration}


\author{Tingfeng Hong}
\orcid{0009-0009-7721-5716}
\affiliation{
  \institution{Kuaishou Technology}
  \city{Beijing}
  \country{China}}
\email{hongtingfeng03@kuaishou.com}
\authornote{Both authors contributed equally to this research.}

\author{Pingye Ren}
\orcid{0009-0001-9766-8140}
\affiliation{
  \institution{Kuaishou Technology}
  \city{Beijing}
  \country{China}}
\email{23210720232@m.fudan.edu.cn}
\authornotemark[1]

\author{Xinlong Xiao}
\orcid{0009-0006-5225-6237}
\affiliation{
  \institution{Kuaishou Technology}
  \city{Beijing}
  \country{China}}
\email{xiaoxinlong@kuaishou.com}

\author{Chao Wang}
\orcid{0009-0000-5516-0771}
\affiliation{
  \institution{Kuaishou Technology}
  \city{Beijing}
  \country{China}}
\email{wang_chao@alumni.sjtu.edu.cn}

\author{Chenyi Lei}
\orcid{0000-0001-6287-3673}
\affiliation{
  \institution{Kuaishou Technology}
  \city{Beijing}
  \country{China}}
\email{leichy@mail.ustc.edu.cn}

\author{Wenwu Ou}
\affiliation{
  \institution{Kuaishou Technology}
  \city{Beijing}
  \country{China}}
\email{ouwenweu@gmail.com}

\author{Han Li}
\affiliation{
  \institution{Kuaishou Technology}
  \city{Beijing}
  \country{China}}
\email{lihan08@kuaishou.com}


\begin{abstract}
Balancing multiple objectives is critical for user satisfaction in modern recommender and search systems, yet current Multi-Task Fusion (MTF) methods rely on static, manually-tuned weights that fail to capture individual user intent. While Reinforcement Learning (RL) offers a path to personalization, traditional approaches often falter due to training instability and the sparse rewards inherent in these large-scale systems. To address these limitations, we propose Group-relative Reinforcement learning with Adaptive Dirichlet Exploration (GRADE), a novel and robust framework for personalized multi-task fusion. GRADE leverages a critic-free, Group Relative Policy Optimization (GRPO) paradigm, enabling stable and efficient policy learning by evaluating the relative performance of candidate weight groups. Its core innovations include employing the Dirichlet distribution for principled and structured exploration of the weight space, and a composite reward function that combines sparse user feedback with dense model priors and rule-based constraints to guide the search effectively. Deployed in the in-app marketplace of an application with over hundreds of millions daily active users, GRADE significantly outperforms established baselines, achieving substantial gains in rigorous large-scale A/B tests: +0.595\% in CTR, +1.193\% in CVR, +1.788\% in OPM, and +1.568\% in total order volume. Following its strong performance, GRADE has been fully deployed in the marketplace search scenario of Kuaishou, serving hundreds of millions of users.
\end{abstract}




\maketitle

\section{Introduction}

\begin{figure*}[ht]
    \centering 
    \includegraphics[width=0.9\textwidth]{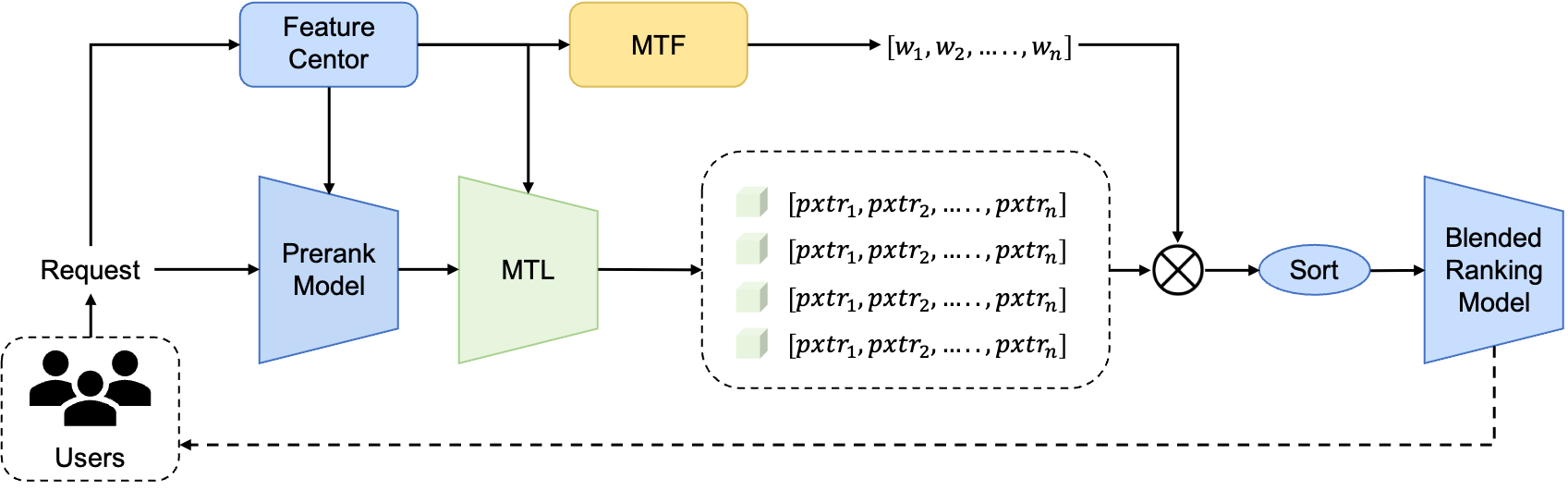} 
    \caption{Overall architecture of the personalized multi-objective ranking system. It comprises: (1) a Feature Center and Prerank Model for initial feature processing and candidate generation; (2) a Multi-Task Learning (MTL) model predicting various user feedback signals; (3) a Multi-Task Fusion (MTF) module (our proposed GRADE framework) that learns personalized weights ($w_1, \dots, w_n$); these weights are then applied to calculate final scores and sorted to generate a blended ranking by the Blended Ranking Model, which ultimately delivers results to users.}
    \label{fig1}
\end{figure*}

A central challenge for modern large-scale internet platforms is to effectively harness diverse user feedback signals across various domains, including video \cite{r13,r14,r15}, live streaming \cite{r16,r17}, e-commerce \cite{r18,r19,r20}, and news dissemination \cite{r21}. Within the e-commerce service of a content platform, for instance, a user's journey generates a multitude of behavioral signals—such as clicks, purchases, and transaction amounts. Enhancing overall user satisfaction is paramount for the vitality of the platform's ecosystem and its commercial viability. Consequently, it is imperative for recommendation systems to balance these multiple, often conflicting, objectives, rather than optimizing for any single metric in isolation.

To achieve this, a two-stage architecture is commonly employed as shown in Figure\ref{fig1}. The first stage utilizes a Multi-Task Learning (MTL) model to simultaneously predict user responses across a spectrum of key performance indicators\cite{r23,r24}, such as Click-Through Rate (CTR), Conversion Rate (CVR), Orders Per Mille (OPM), and Gross Merchandise value Per Mille (GPM). Subsequently, a Multi-Task Fusion (MTF) module synthesizes these disparate predictions, computing a single, coherent score for each item that ultimately determines the final ranking presented to the user.

While this modular design is effective, the performance of the entire recommendation pipeline often hinges on the intelligence of the MTF module. The prevailing industry practice is to implement this fusion logic using pre-defined, static formulas, typically a weighted sum or product of the MTL outputs. This legacy approach, however, presents significant obstacles to further progress. Operationally, it locks engineers into a slow and laborious cycle of manual weight tuning, followed by extensive A/B testing to validate changes. More fundamentally, this undifferentiated approach is conceptually flawed for personalization. A single, global weighting scheme cannot dynamically adapt to the unique intentions of each user, which can shift even within a single session. In e--commerce, for instance, a user often transitions from a broad exploration phase to a focused purchasing phase, each warranting a different balance of objectives (e.g., maximizing clicks versus conversions). A static model is incapable of capturing this nuance, leading to a suboptimal and generic user experience.

To overcome the limitations of static fusion, Reinforcement Learning (RL) has emerged as a promising paradigm for developing dynamic and personalized weighting strategies \cite{r25,r26, batchrl,rlcw,rlvw,UnifiedRL}. Conceptually, RL is well-suited for this task, as it enables a recommendation system to continuously adapt to evolving user preferences by maximizing an overall reward from environmental feedback. However, its application to large-scale industrial scenarios is fraught with challenges. The most prevalent approaches, based on Policy Gradient and Actor-Critic architectures, introduce significant hurdles for deployment: the complex structure presents challenges for stable training with sparse production rewards, and the exploration mechanisms are often inefficient for the high-dimensional, continuous space of fusion weights. These intertwined challenges of complexity, instability, and inefficient exploration motivate our search for a simpler, more direct, and robust optimization framework.

To address the challenges of model complexity, reward sparsity, and inefficient exploration, our approach draws inspiration from Group Relative Policy Optimization (GRPO) \cite{r22}. The core of GRPO is that it learns from the relative preferences within a group of candidate outputs, rather than relying on a critic's estimate of absolute value. This paradigm has proven highly effective for fine-tuning large language models (LLMs), particularly in complex reasoning tasks such as mathematics.  We pos it a strong parallel to our fusion task: from a post-hoc perspective, a user's explicit feedback (e.g., a click or purchase) reveals a "ground-truth" preference, indicating which combination of objective weights would have been optimal. This insight makes a relative preference-based learning framework like GRPO an ideal choice.

Based on this foundation, we propose GRADE (Group-relative Reinforcement learning with Adaptive Dirichlet Exploration), a novel framework that operationalizes these ideas. To ensure a stable and effective learning process, GRADE is trained in a two-stage paradigm: an initial supervised learning-to-rank phase provides a robust baseline policy, which is then fine-tuned using our GRPO-based framework centered on three core innovations. First, it adapts the critic-free GRPO paradigm, directly addressing the instability of conventional Actor-Critic methods. By learning from the relative quality of multiple candidate weight vectors simultaneously, it obviates the need for an unstable critic network while also providing powerful, inherent exploration capabilities through its intra-group comparison. Second, for policy exploration in the continuous weight space, we employ the Dirichlet distribution. Unlike unconstrained distributions like the Gaussian which require a subsequent normalization step, the Dirichlet distribution inherently generates samples that satisfy the non-negative and sum-to-one constraints of fusion weights. This property, combined with its efficiency in high dimensions and controllable stochasticity, makes it a superior choice for this task. Third, we design a composite reward function to provide a rich and stable learning signal, combining posterior rewards from user feedback, dense prior rewards from model predictions to mitigate sparsity, and rule-based format rewards that incorporate human heuristics to prevent weight polarization and resolve reward homogenization.

\begin{figure*}[ht]
    \centering 
    \includegraphics[width=0.9\textwidth]{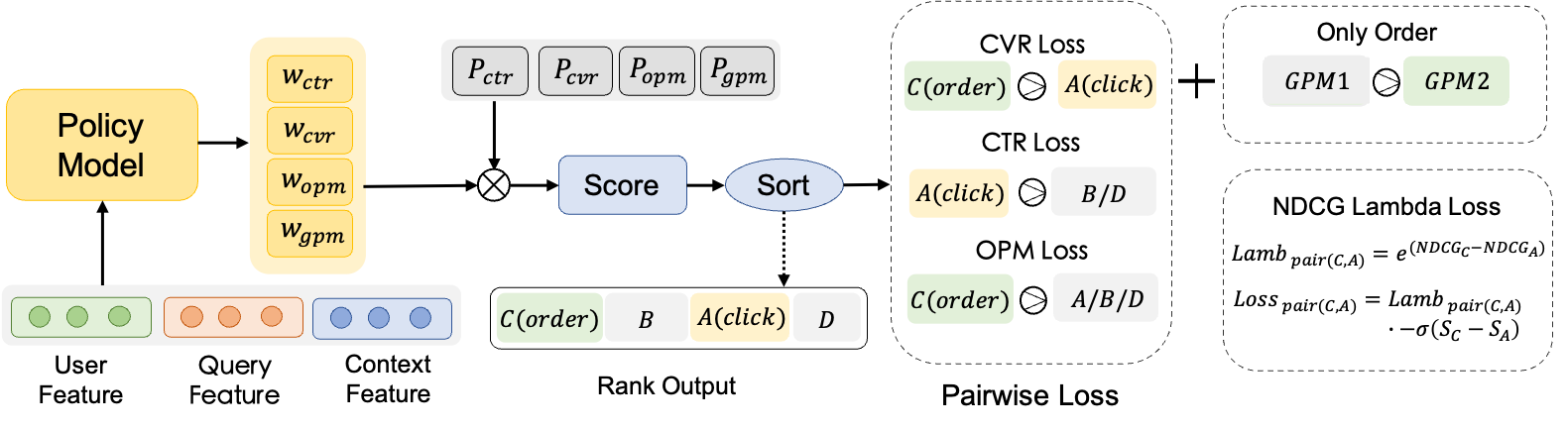} 
    \caption{The architecture of the Stage 1 supervised Learning-to-Rank (LTR) model, which is pre-trained to provide a robust baseline policy. The model learns to generate fusion weights using a multi-objective pairwise loss based on the LambdaLoss framework. In the depicted LambdaLoss formula, $S_c$ and $S_a$ represent the fusion scores of two items, and $\sigma$ denotes the sigmoid function.} 
    \label{fig3}
\end{figure*}

The primary contributions of this work are as follows:
\begin{itemize}
    \item We adapt the critic-free Group Relative Policy Optimization (GRPO) paradigm to enable stable and efficient policy learning for personalized multi-task fusion.
    \item We propose a principled exploration strategy using the Dirichlet distribution, which is uniquely suited for the constrained, high-dimensional space of fusion weights.
    \item We design a multi-faceted composite reward function that combines posterior, prior, and heuristic signals to provide robust policy guidance and prevent reward hacking.
\end{itemize}

\section{Related Work}

\subsection{Multi-Task Learning}

In contrast to Single-Task Learning (STL), where tasks are modeled in isolation, Multi-Task Learning (MTL) offers a paradigm that jointly trains multiple related tasks within a single model. The foundational assumption is that leveraging shared information across tasks acts as an inductive bias to enhance the model's overall generalization performance. The primary distinction among MTL architectures lies in their parameter sharing strategy. The most common approach is hard parameter sharing, which employs a shared feature encoder across all tasks, with separate task-specific output layers branching off from this common base \cite{r4}. A more flexible alternative is soft parameter sharing, where each task maintains its own parameters, but these are encouraged to be similar through mechanisms like regularization or the dynamic gating found in Mixture-of-Experts (MoE) models \cite{r5,r6}. More sophisticated architectures, such as Progressive Layered Extraction (PLE), employ customized sharing mechanisms that use a collection of "expert" networks, some of which are shared to learn common patterns while others are dedicated to specific tasks to capture unique features \cite{r7,r8}.


\subsection{Multi-task Fusion}
Multi-Task Fusion (MTF) constitutes the critical final stage in a multi-objective recommendation pipeline, where various predicted scores are integrated into a single value for item ranking. The core challenge of MTF is to enhance a user's overall satisfaction—a latent and composite construct. Unlike MTL tasks that often have discrete labels, this holistic satisfaction is an unlabeled objective that must be inferred from multiple behavioral signals within a session, such as clicks, conversions, and transaction amounts. Conventional approaches typically rely on a predefined fusion formula (e.g., a weighted sum) and employ black-box optimization techniques like grid search\cite{gridsearch} or Bayesian optimization\cite{beiyesi, r11} to find a single, globally optimal set of coefficients. Furthermore, to address challenges with offline-online consistency, methods like Evolution Strategies (ES)\cite{es} have been employed for direct online exploration of fusion weights. However, these methods are inherently non-personalized, applying a uniform strategy to all users and thus failing to capture individual preferences.

To address this lack of personalization, the field has increasingly turned to Reinforcement Learning (RL) to dynamically optimize the fusion weights. In this formulation, the problem is modeled by treating the set of fusion weights as actions taken by the agent. The recommender system, along with each user's personalized context, constitutes the environment, and the subsequent user feedback (e.g., clicks, conversions) serves as the reward signal. By learning a policy that maps the environment state to optimal actions, this approach aims to find personalized fusion weights that maximize the expected reward for each user, ultimately maximizing business value. 

For safe deployment in large-scale systems, RL-based MTF is often implemented in an offline (batch) setting, learning from historical logs. The primary challenge in this setting is the distributional shift between the learned policy and the logged data, which can lead to significant extrapolation errors. To mitigate this, many approaches adapt foundational offline RL algorithms designed to constrain the policy against out-of-distribution actions. Notable examples of this paradigm include Batch-Constrained Deep Q-Learning (BCQ) \cite{bcq}, Conservative Q-Learning (CQL) \cite{cql}, and specific MTF applications like BatchRL-MTF \cite{batchrl}.

Another prevalent branch of research within this domain utilizes Policy Gradient and Actor-Critic methods. Foundational algorithms like the Deep Deterministic Policy Gradient (DDPG)\cite{ddpg} established a robust actor-critic framework for the continuous action spaces inherent in MTF. Building on this paradigm, subsequent works have tailored it for specific recommendation scenarios. For instance, RLUR\cite{RLUR} prioritizes user retention as its primary modeling objective based on long-term value considerations. UNEX-RL\cite{unex} is proposed to solve the observation dependency and cascading effect problems in multi-stage recommender systems. UnifiedRL\cite{UnifiedRL} attempts to improve exploration efficiency by implementing a multi-round strategy of online exploration using clipped Gaussian perturbations, followed by offline training. Despite its sophistication, this approach is limited as it explores only a single weight combination at a time, restricting the breadth of the search.

While this Actor-Critic paradigm has proven powerful, it comes with fundamental limitations. These frameworks rely on complex architectures that are notoriously difficult to train and tune stably, particularly with the sparse reward signals in e-commerce. Furthermore, their exploration strategies are often inefficient at searching the high-dimensional, continuous space of fusion weights, leading to slow convergence and suboptimal policies. These critical challenges—the inherent complexity and instability of mainstream RL models, coupled with inefficient exploration—underscore the need for a simpler, more robust framework. This motivates the design of our proposed GRADE framework, which circumvents these issues by a more direct and efficient approach to personalized parameter optimization.

\subsection{The GRPO Paradigm}
Group Relative Policy Optimization (GRPO) is an advanced reinforcement learning algorithm originating from the fine-tuning of large language models (LLMs) \cite{r22}. It enhances traditional policy gradient methods by shifting the learning paradigm from estimating the absolute value of a single action to learning from the relative preferences among a group of candidate actions. This approach is particularly effective in tasks where an optimal output exists but is hard to discover directly.

The GRPO objective function, which the policy network $\pi_{\theta}$ aims to maximize, is typically formulated as follows:
\begin{equation}
\mathcal{J}_{GRPO}(\theta) = \mathbb{E}_{s \sim \mathcal{D}, \{a_i\}_{i=1}^k \sim \pi_{\theta}(s)} \left[ \sum_{i=1}^k \frac{\pi_{\theta}(a_i|s)}{\pi_{\theta_{old}}(a_i|s)} \hat{A}_i \right].
\end{equation}
Here, for a given state $s$ (e.g., a user request), the policy $\pi_{\theta}$ generates a group of $k$ actions $\{a_i\}_{i=1}^k$ (e.g., candidate weight vectors). $\pi_{\theta_{old}}$ is the policy from the previous iteration, used for importance sampling. The core of the method lies in the calculation of the normalized relative advantage, $\hat{A}_i$, for each action $a_i$. This is typically computed by first obtaining a reward $r_i$ for each action from a reward model, and then normalizing these rewards within the group: 
\begin{equation}
\hat{A}_i = (r_i - \mu_r) / (\sigma_r + \epsilon),
\end{equation}
where $\mu_r$ and $\sigma_r$ are the mean and standard deviation of rewards in the group. The policy is then updated to increase the probability of actions with a positive relative advantage.

This group-wise comparison mechanism endows GRPO with several key advantages:
\begin{itemize}
    \item \textbf{Avoidance of a Critic Network:} Its most notable advantage is bypassing the need for a separate Critic network to estimate a state-value function. This makes the framework simpler and more stable, especially in scenarios with sparse or noisy rewards.
    \item \textbf{Enhanced Data Efficiency:} By learning from a ranked set of $k$ outputs simultaneously, GRPO extracts a much richer learning signal from a single state transition compared to methods that learn from a single action feedback.
    \item \textbf{Improved Training Stability:} The normalization of rewards into a relative advantage reduces the sensitivity of the training process to the absolute scale and variance of the reward function, leading to smoother and more reliable convergence.
\end{itemize}

The application of GRPO to our MTF task is well-justified by drawing a strong parallel to its successful use in LLM reasoning tasks. In mathematical reasoning, an LLM must find the correct reasoning path among many possibilities. Similarly, in MTF, for any user request, a post-hoc "correct answer" exists in the form of an optimal weight combination that best aligns with the user's subsequent feedback (e.g., click, purchase). GRPO is uniquely suited to find this "answer" by efficiently exploring and comparing groups of candidate weights. Furthermore, it directly addresses the identified weaknesses of traditional RL methods in the MTF context: its Critic-free architecture provides robustness against the sparse rewards common in e-commerce, and its structured, group-wise learning offers a more efficient and stable alternative to prior exploration techniques.

\section{GRADE Framework}

This section elaborates on the architecture and training methodology of the GRADE framework. We begin by describing its two-stage training paradigm, which is designed to ensure both a stable initialization and effective policy exploration. We then delve into the core optimization techniques that power the reinforcement learning (RL) fine-tuning stage. These methods collectively enable GRADE to perform an automated parameter search, learning a policy that dynamically adapts fusion weights to maximize the overall reward.

\begin{figure*}[t]
    \centering 
    \includegraphics[width=0.9\textwidth]{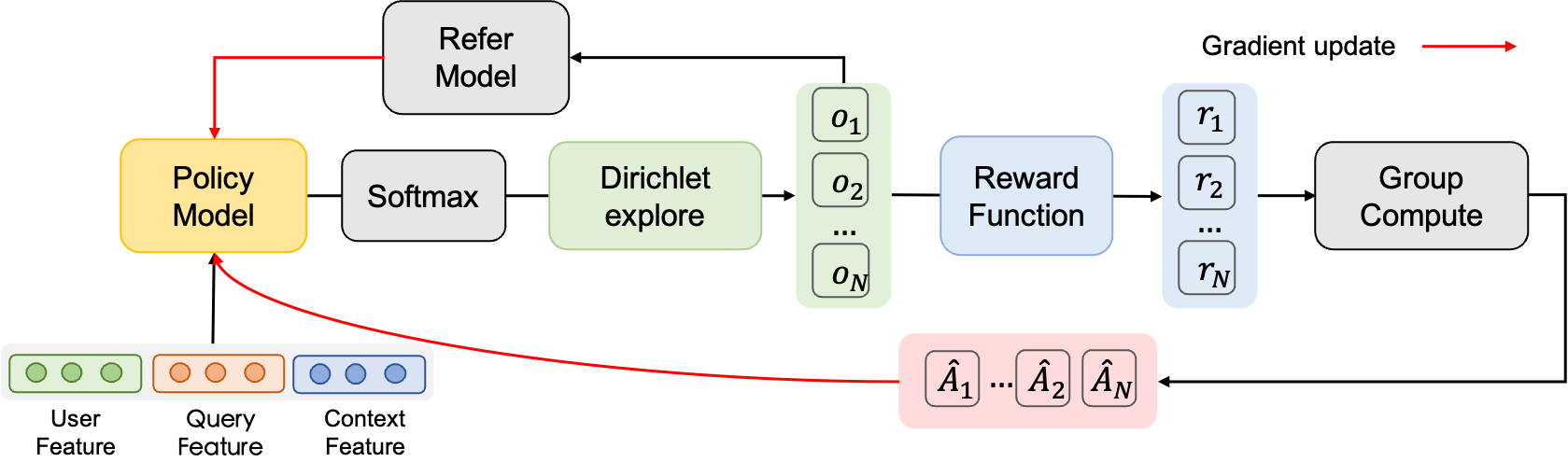} 
    \caption{The training loop for Stage 2: GRPO-based fine-tuning.} 
    \label{fig4}
\end{figure*}

\subsection{Training Paradigm of GRADE}

The training of the GRADE framework is structured as a two-stage paradigm, a design choice intended to strike a balance between stable convergence and effective policy exploration. The objective is to first establish a robust, generalized policy through supervised learning before fine-tuning it for personalized parameter search via reinforcement learning.

The policy model maintains a consistent architecture across both stages. It processes a comprehensive set of input features, including the user's profile and historical behaviors, the query's categorical and semantic attributes, and other contextual signals. Notably, the policy model is item-agnostic; it does not receive any item-side features. The model's output constitutes the agent's action: a vector of fusion weights, $\boldsymbol{w}=[w_1, w_2, \dots, w_n]$, which are then applied to the corresponding predicted scores (pCTR, pCVR, etc.) of each candidate item. For the final fusion, we employ a weighted additive formula, and a softmax function is applied to the model's output layer to ensure the generated weights are normalized and stable.

It is important to clarify that while we use a weighted additive formula for its straightforward interpretability, the GRADE framework itself is agnostic to the specific fusion function. It is broadly applicable to any scenario requiring the dynamic personalization of parameters.

The two stages are detailed as follows:
\begin{enumerate}
    \item \textbf{Stage 1: Supervised Learning-to-Rank (LTR) Initialization.} The initial stage trains the policy network using a supervised LTR objective. The primary goal is to provide a robust and well-grounded initialization for the model, ensuring it aligns with general user preferences before the reinforcement learning phase begins.
    \item \textbf{Stage 2: GRPO-based Reinforcement Learning Fine-tuning.} The second stage leverages the GRPO framework to fine-tune the pre-trained model. In this phase, the model learns to conduct an automated, personalized parameter search, moving beyond the supervised baseline to optimize for the holistic user satisfaction metric.
\end{enumerate}

\subsubsection{Stage 1: Supervised LTR Initialization}
This stage provides a robust initialization for the policy network via a multi-objective Learning-to-Rank (LTR) approach. We employ the LambdaLoss framework \cite{lambdaloss1, lambdaloss2} as show in Figure\ref{fig3}, a pairwise method that intelligently weights training samples by their impact on the final ranking metric. To optimize for multiple objectives simultaneously, the total loss is a weighted summation of individual LambdaLoss terms, each corresponding to a key user feedback metric:
\begin{equation}
\mathbf{L}_{\text{total}} = \alpha_{\text{ctr}}\mathbf{L}_{\lambda}^{\text{ctr}} + \alpha_{\text{cvr}}\mathbf{L}_{\lambda}^{\text{cvr}} + \alpha_{\text{opm}}\mathbf{L}_{\lambda}^{\text{opm}} +
\alpha_{\text{gpm}}\mathbf{L}_{\lambda}^{\text{gpm}}.
\end{equation}
The policy network pre-trained with this objective serves as the reference model for the fine-tuning stage.

\subsubsection{Stage 2: GRPO-based Reinforcement Learning Fine-tuning}
The second stage fine-tunes the pre-trained model from Stage 1 using the GRPO reinforcement learning framework, as depicted in Figure \ref{fig4}. The goal is to move beyond the generalized supervised policy to discover personalized, high-performance weighting strategies. This stage involves two models: a trainable Policy Model ($\pi_{\theta}$) and the frozen Reference Model ($\pi_{\text{ref}}$), which is a stable snapshot from a previous training iteration. At the beginning of each daily training iteration, the policy model is synchronized with the reference model ($\theta \leftarrow \theta_{\text{ref}}$), and this initial policy is denoted as the old policy ($\pi_{\theta_{\text{old}}}$) for data collection.

A key distinction from typical RL applications in LLMs is our continuous action space. For exploration, we construct a stochastic policy by sampling actions from a probability distribution $\mathcal{D}$, which is parameterized by the deterministic output of the policy network, $\boldsymbol{\pi}_{\theta}(q)$. A stochastic action $o$ (a specific weight vector) is thus sampled as:
\begin{equation}
    o \sim \mathcal{D}(\boldsymbol{\pi}_{\theta}(q)).
\end{equation}
The training process proceeds as follows. For a given state $q$, we sample a group of $G$ actions $\{o_i\}_{i=1}^G$ from the old policy $\pi_{\theta_{\text{old}}}$. Each action $o_i$ is evaluated to obtain a reward $r_i$ (detailed in Section 3.2.1). From the group's rewards, we then compute the normalized relative advantage $\hat{A}_i$ for each action:
\begin{equation}
    \hat{A}_i = \frac{r_i - \text{mean}(\{r_j\}_{j=1}^{G})}{\text{std}(\{r_j\}_{j=1}^{G})}.
\end{equation}
The policy model $\pi_{\theta}$ is subsequently updated by maximizing the GRPO objective function:
\begin{equation}
\begin{split}
\mathcal{J}_{\text{GRPO}}(\theta) = \mathbb{E}_{q, \{o_i\} \sim \pi_{\theta_{\text{old}}}} \Biggl[ & \frac{1}{G} \sum_{i=1}^G \min \Bigl( r_i(\theta) \hat{A}_{i} , \\
& \qquad \text{clip} \left( r_i(\theta), 1 - \epsilon, 1 + \epsilon \right) \hat{A}_{i} \Bigr) \Biggr] \\
& - \rho D_{\text{KL}}[\pi_{\theta} || \pi_{\text{ref}}],
\end{split}
\end{equation}
where the importance sampling ratio $r_i(\theta)$ is the ratio of probabilities of action $o_i$ under the new and old policies:
\begin{equation}
    r_i(\theta) = \frac{p(o_i | \mathcal{D}(\pi_{\theta}(q)))}{p(o_i | \mathcal{D}(\pi_{\theta_{\text{old}}}(q)))}.
\end{equation}
Here, $p(\cdot|\mathcal{D})$ denotes the probability density function of the chosen exploration distribution $\mathcal{D}$. This objective function incorporates two key stabilization mechanisms. The PPO-style clip function constrains the magnitude of the policy update to prevent destructive steps. Additionally, the KL-divergence penalty, $D_{\text{KL}}[\pi_{\theta} || \pi_{\text{ref}}]$, regularizes the updated policy to stay close to the original supervised policy $\pi_{\text{ref}}$, mitigating catastrophic forgetting. Together, these components ensure stable and constrained policy updates during fine-tuning.

\subsection{Policy Optimization of GRADE}
While the core GRPO paradigm is effective, its practical application required further optimization to address two key challenges observed during development. First, we found that naive exploration strategies for the continuous weight space, such as grid sampling, were computationally inefficient. Second, relying on a simple performance metric as the reward signal led to "reward hacking." This phenomenon, where the policy over-optimizes for the training data, often results in out-of-distribution (OOD). To overcome these issues, we introduce two critical enhancements to the GRADE framework: 1) a principled exploration mechanism using Dirichlet distribution sampling, and 2) a composite, rule-based reward function to guide the policy towards more robust solutions.

\subsubsection{Dirichlet Distribution for Structured Exploration}
In the RL fine-tuning stage, a critical challenge is designing an effective exploration strategy for the continuous action space of fusion weights. This requires a probability distribution that can intelligently sample new weight candidates. While the Gaussian distribution is a standard choice for continuous spaces, we posit that it is suboptimal for this task. Instead, we propose the use of the Dirichlet distribution as a more principled and efficient exploration mechanism for several key reasons.

First, the Dirichlet distribution inherently respects the structural constraints of fusion weights. These weights must be non-negative and sum to one (i.e., lie on the probability simplex). The Dirichlet distribution naturally generates samples that meet these criteria in a single step, whereas a Gaussian distribution would require a cumbersome two-step process: sampling from an unconstrained space and then projecting the samples onto the simplex using a potentially distorting function like Softmax.

Second, it offers intuitive control over the exploration-exploitation trade-off via a single concentration parameter, $\hat{\alpha}$. A smaller $\hat{\alpha}$ yields a more uniform (exploratory) distribution, while a larger $\hat{\alpha}$ leads to samples highly concentrated around the mean (exploitative). This parameter can be easily annealed during training to gradually transition the policy from exploration to exploitation.

Finally, it enables targeted exploration centered around the learned policy. We construct the Dirichlet's parameters, $\boldsymbol{\alpha}$, by scaling the policy network's deterministic output, $\boldsymbol{\pi}_{\theta}(q)$, with the total concentration $\hat{\alpha}$:
\begin{equation}
    \boldsymbol{\alpha} = \hat{\alpha} \cdot \boldsymbol{\pi}_{\theta}(q).
\end{equation}
A key property of the Dirichlet distribution is that the expected value of a sampled vector is identical to the normalized parameter vector. This ensures that our exploration is not random but is intelligently centered around the policy's current recommendation:
\begin{equation}
    \mathbb{E}[\mathbf{p}] = \frac{\boldsymbol{\alpha}}{\sum \alpha_i} = \frac{\hat{\alpha} \cdot \boldsymbol{\pi}_{\theta}(q)}{\hat{\alpha} \sum \pi_i} = \boldsymbol{\pi}_{\theta}(q),
\end{equation}
which leads to a significantly more efficient search process.

The probability density function (PDF) for the Dirichlet distribution is given by:
\begin{equation}
\text{Dir}(\mathbf{p} \mid \boldsymbol{\alpha}) = \frac{\Gamma\left(\sum_{i=1}^{k} \alpha_i\right)}{\prod_{i=1}^{k} \Gamma(\alpha_i)} \prod_{i=1}^{k} p_i^{\alpha_i - 1} ,
\end{equation}
where $\Gamma(z)$ is the gamma function:
\begin{equation}
\Gamma(z) = \int_{0}^{\infty} t^{z-1} e^{-t} \, dt.
\end{equation}
Collectively, these properties make the Dirichlet distribution a superior and highly suitable choice for the exploration mechanism within our GRADE framework.

\subsubsection{Composite Reward Function}
To effectively guide the RL agent towards an optimal policy, we design a composite reward function that balances three objectives: direct performance optimization, mitigation of signal sparsity, and the incorporation of human priors. The total reward, $R_{\text{total}}$, for a given weight vector is a weighted sum of three distinct components:
\begin{equation}
    R_{\text{total}} = \lambda_1 R_{\text{post}} + \lambda_2 R_{\text{prior}} + \lambda_3 R_{\text{format}},
\end{equation}
where $\lambda_1, \lambda_2, \lambda_3$ are hyperparameters that control the contribution of each component.

\paragraph{Posterior Reward ($R_{\text{post}}$)}
This primary reward component directly measures policy performance based on actual user feedback. For a given ranked list, we calculate the Normalized Discounted Cumulative Gain (NDCG). It is defined as the ratio of the Discounted Cumulative Gain (DCG) of the ranked list to the Ideal DCG (IDCG), which is the DCG of a perfectly sorted list:
\begin{equation}
    \text{NDCG} = \frac{\text{DCG}}{\text{IDCG}},
\end{equation}
where the DCG is calculated by summing the relevance scores of items, logarithmically discounted by their rank $i$:
\begin{equation}
    \text{DCG} = \sum_{i=1}^{N} \frac{2^{rel_i} - 1}{\log_2(i+1)},
\end{equation}
where $N$ is the number of items in the list, $rel_i$ is the ground-truth relevance score of the item at rank $i$, and IDCG is the DCG score of the ideal ranking. We compute this metric for multiple key objectives. For objectives like CTR, CVR, and OPM, the relevance score $rel_i$ is a binary value (e.g., 1 for a click or purchase, 0 otherwise). For Gross Merchandise Value (GPM), we use the predicted score (pGPM) as a graded relevance signal, allowing the policy to optimize for expected value. The total posterior reward is the weighted sum of these individual NDCG scores.

\paragraph{Prior Knowledge Reward ($R_{\text{prior}}$)}
A key challenge in our domain is the extreme sparsity of posterior signals like conversions, which can impede learning. To provide a denser learning signal, we introduce a reward based on the prior knowledge captured in the MTL model's dense predictions (e.g., pCTR, pCVR). The underlying principle is that ranking items with high predicted scores higher should correlate with better online performance. This reward is also formulated as an NDCG, using the predicted scores as the relevance signals.

\paragraph{Weight Format Reward ($R_{\text{format}}$)}
Weight Format Reward acts as a conditional regularization mechanism to guide the search process and incorporate domain knowledge. It is only triggered when the sum of the primary rewards ($R_{\text{post}} + R_{\text{prior}}$) is positive, ensuring it refines already promising candidates, which is implemented via a gating function, $F_g(\cdot)$, that evaluates the relative magnitudes of the weights. The input to this function is the difference between a target weight and a specified proportion of other weights, as illustrated by the following examples:
\begin{equation}
R_{\text{opm}} = F_{g}\left(w_{opm} - \alpha\% \cdot \max(w_{ctr}, w_{cvr}, w_{gpm})\right),
\end{equation}
\begin{equation}
R_{\text{cvr}} = F_{g}\left(w_{cvr} - \beta\% \cdot \max(w_{ctr}, w_{gpm})\right).
\end{equation}
The gating function $F_g(x)$ is a piecewise function designed to create a "soft constraint" on the weight structure, defined as:
\begin{equation}
F_{\text{g}}(x) = 
\begin{cases} 
e^x - 1, & x < 0 \\
\frac{\tau}{\pi} \sin\left(\frac{\pi x}{\tau}\right), & x \in [0, \tau] \\
e^{-x+\tau} - 1, & x > \tau 
\end{cases} ,
\end{equation}
where $\tau$ is a hyperparameter controlling the width of the positive reward region. The behavior of this function is three-fold. For negative inputs ($x < 0$), a penalty is applied, which discourages weights from being smaller than their desired proportion. For inputs in a small positive range ($x \in [0, \tau]$), it provides a positive reward, defining a desirable range that encourages a specific hierarchical structure. For inputs exceeding this range ($x > \tau$), a penalty is applied, thus restricting the dominance of excessive weight. In essence, $F_g(x)$ encodes our heuristic preference for a balanced yet structured weight distribution. In our implementation, we set $\tau = 0.4$. It offers three key benefits:
\begin{enumerate}
    \item \textbf{Injecting Human Priors:} It allows us to encode heuristic knowledge about what constitutes a "good" weight distribution, such as ensuring a primary business objective like OPM maintains a dominant weight.
    \item \textbf{Preventing Polarization:} It penalizes extreme weight distributions, encouraging a more balanced consideration of all objectives. While conceptually similar to the hard constraints in methods like BCQ\cite{bcq} and UnifiedRL\cite{UnifiedRL}, our use of a soft constraint preserves greater exploratory flexibility.
    \item \textbf{Breaking Ties:} In later training stages, when multiple weight combinations yield identical rewards, this component acts as a tie-breaker by favoring the most well-structured vector, thus preventing policy stagnation.
\end{enumerate}

\section{Experiment}
\subsection{Dataset and Evaluation Metrics}
Our experiments are conducted on a large-scale dataset comprising 1.8 billion user search sessions, collected from the e-commerce search service of Kuaishou, of which 10\% is held out for testing. Each session instance contains a user's query, the exposed item list, and corresponding feedback signals such as clicks and purchases. Model performance is assessed using a dual offline and online evaluation protocol. The individual $NDCG_{xtr}$ score for each session is computed using interactions like clicks and purchases as binary relevance signals. A specific strategy is used for the $NDCG_{GPM}$ metric: the top two items by pGPM in any converted session are used as positive labels to specifically evaluate high-value item ranking. For online validation, models are deployed in a rigorous A/B testing environment, where their real-world efficacy is measured by the impact on key business metrics.

\subsection{Experimental Settings} \label{setting}
The GRADE policy network outputs a four-dimensional weight vector, $\boldsymbol{w} = [w_{ctr}, w_{cvr}, w_{opm}, w_{gpm}]$, used to fuse the predicted scores. The network employs an MMoE-like\cite{r5} architecture and is trained using the Adam optimizer with a batch size of 2048. For the GRPO update, we sample a group of $G=20$ candidate actions for each state. The concentration parameter $\hat{\alpha}$ of the Dirichlet distribution follows a cosine annealing schedule, cycling between a minimum of 5 and a maximum of 15 every 50000 iterations. The coefficients for the posterior, prior, and format rewards are set to $\lambda_1=1.0$, $\lambda_2=0.3$, and $\lambda_3=0.05$, respectively. It is important to note that these hyperparameters were tuned for our specific application; transferring GRADE to other scenarios would likely require re-tuning.

\begin{table}[h]
  \caption{Offline Performance Comparison (NDCG scores)}
  \label{tab:offline}
  \centering
  \begin{tabular}{lcccc}
    \toprule
    Method & CTR & CVR & CTCVR & GPM \\
    \midrule
    Formula & 0.607 & 0.772 & 0.686 & 0.892 \\
    SP      & \textbf{0.632} & 0.774 & 0.689 & 0.886 \\
    BCQ     & 0.620 & 0.777 & 0.691 & 0.894 \\
    DDPG    & 0.624 & 0.779 & 0.693 & 0.893 \\
    GRADE   & 0.625 & \textbf{0.782} & \textbf{0.697} & \textbf{0.895} \\
    \bottomrule
  \end{tabular}
\end{table}

\subsection{Offline Evaluation}
We compare GRADE against four representative baseline methods:
\begin{itemize}
    \item \textbf{Formula:} The fixed-weight fusion formula, which incorporates coarse-grained personalization for different user populations.
    \item \textbf{BCQ:} A BCQ\cite{bcq} based MTF agent, adapted for our scenario referencing recent industrial work \cite{batchrl}.
    \item \textbf{DDPG:} A DDPG \cite{ddpg} based agent, informed by related works \cite{UnifiedRL, RLUR} to ensure it is a robust competitor.
    \item \textbf{Supervised Policy (SP):} The policy network trained only with the supervised LTR objective from Stage 1, serving as an ablation to measure the contribution of RL fine-tuning.
\end{itemize}
For a fair comparison, all neural network-based models are built upon the same MMoE-like architecture as GRADE and undergo a thorough hyperparameter search to achieve their best performance.
The offline results in Table \ref{tab:offline} demonstrate that GRADE outperforms all baselines on key business-related metrics. While the supervised baseline (SP) achieves the highest $NDCG_{CTR}$ by effectively optimizing for local pairwise signals, it fails to capture the list-wise optimality required for more complex conversion-based objectives. Other RL methods like BCQ and DDPG are constrained by less efficient exploration, leading to suboptimal policies. In contrast, GRADE's robust learning paradigm and structured exploration allow it to find a superior policy that better balances the full spectrum of user feedback signals.

\begin{table}[h]
  \caption{Online A/B Test Results}
  \label{tab:online}
  \centering
  \begin{tabular}{lcccc}
    \toprule
    Method & CTR & CVR & OPM & GPM \\
    \midrule
    Formula & - & - & - & - \\
    SP      & \textbf{+0.73\%} & +0.03\% & +0.56\% & +0.08\%   \\
    GRADE   & +0.60\% & \textbf{+1.19\%} & \textbf{+1.78\%} &\textbf{+0.52\%} \\
    \bottomrule
  \end{tabular}
\end{table}

\begin{table}[h]
  \caption{Ablation study on exploration intensity (NDCG scores)}
  \label{tab:abla-offline}
  \centering
  \begin{tabular}{ccccc}
    \toprule
    Group Size ($G$) & Concentration ($\hat{\alpha}$) & CTR & CVR & OPM \\
    \midrule
    5 & $C$  & \textbf{0.636} & 0.775 & 0.688 \\
    10 & $C$  & 0.632 & 0.778 & 0.691 \\
    20 & 5  & 0.622 & 0.781 & 0.695 \\
    20 & 10  & 0.630 & 0.779 & 0.694 \\
    20 & 15 & 0.632 & 0.778 & 0.691 \\
    20 & $C$  & 0.625 & \textbf{0.782} & \textbf{0.697} \\
    30 & $C$  & 0.624 & 0.781 & 0.695 \\
    40 & $C$  & 0.626 & 0.780 & 0.696 \\
    \bottomrule
  \end{tabular}
\end{table}

\subsection{Online Evaluation}
Table \ref{tab:online} summarizes the results of a large-scale two-week A / B test, measuring the relative improvements against the production baseline. The Supervised Policy achieved a strong gain in CTR (+0.73\%) but showed only marginal impact on conversion metrics. In contrast, GRADE delivered substantial improvements across all key business metrics, achieving significant gains in CVR (+1.19\%) and OPM (+1.78\%). The improvements for CTR, CVR, and OPM were all statistically significant (p < 0.05). The superior and more balanced performance on critical conversion and revenue-related metrics validated the effectiveness of the GRADE framework. Consequently, GRADE has been fully deployed in the marketplace search scenario of Kuaishou.

\begin{table}[h]
  \caption{Ablation Study on Reward Components}
  \label{tab:reward_ablation}
  \centering
  \begin{tabular}{lcccc}
    \toprule
    Method & CTR & CVR & OPM & GPM \\
    \midrule
    GRADE ($R_{post}$) & \textbf{+1.04\%} & +0.55\% & +0.93\% & -0.46\%    \\
    GRADE (Full) & +0.60\% & \textbf{+1.19\%} & \textbf{+1.78\%} & \textbf{+0.52\%} \\
    \bottomrule
  \end{tabular}
\end{table}

\subsection{Ablation Study}
The performance of GRADE is highly sensitive to the exploration strategy, in which the exploration intensity is mainly governed by the group size $G$ for sampling and the Dirichlet distribution's concentration parameter $\hat{\alpha}$. Table \ref{tab:abla-offline} presents our ablation study on these parameters, where $C$ denotes the cosine annealing strategy for $\hat{\alpha}$ as described in \ref{setting}.

The results show that as the group size $G$ increases, the performance on key business metrics improves, reaching its maximum at $G=20$. Performance does not monotonically increase with group size; we hypothesize this is because excessively large groups can lead to reward homogenization, diminishing the quality of the relative advantage signal. The experiments on a fixed concentration $\hat{\alpha}$ (with $G=20$) further reveal that lower, more explorative values are preferable. Crucially, the cosine annealing strategy, which dynamically balances exploration and exploitation, achieves the best overall performance on the most critical business metrics. This underscores that a well-calibrated, dynamic exploration strategy is paramount for GRADE's success, rather than simply maximizing the quantity of exploration.

To validate our composite reward function, we conducted an online A/B test comparing the full GRADE model against an ablated version that uses only the posterior reward ($R_{post}$), with results presented in Table \ref{tab:reward_ablation}. The ablated model achieved a slightly higher CTR but performed significantly worse on all other business metrics, even causing a negative impact on GPM. In contrast, the full GRADE model, while trading a minor amount of CTR, delivered substantial gains across all conversion and revenue-related metrics. This result confirms that the prior and format rewards are critical components that regularize the policy, preventing it from overfitting to sparse, direct-engagement signals (i.e., reward hacking) and guiding it towards a more robust, balanced strategy that aligns with overall business objectives.

\section{Conclusion}
In this work, we first identify the critical limitations of conventional RL methods for multi-task fusion, namely the instability of Actor-Critic architectures with sparse rewards and inefficient exploration. In response, we propose GRADE, a robust and efficient framework for personalized fusion in large-scale recommender systems. GRADE adapts the critic-free Group Relative Policy Optimization (GRPO) paradigm, enabling stable policy learning by evaluating relative preferences among candidate actions, thereby obviating the need for an unstable critic. The framework is further enhanced by a principled exploration strategy using the Dirichlet distribution and a composite reward function that mitigates sparsity and prevents reward hacking. Extensive offline and online experiments have demonstrated its effectiveness. Given its strong performance and stability, GRADE has been fully deployed in our production environment, serving hundreds of millions of users daily.

\newpage

\bibliographystyle{ACM-Reference-Format}
\bibliography{sample-base}

\end{document}